\documentclass[letterpaper]{article} 
\usepackage{aaai25}  
\usepackage{times}  
\usepackage{helvet}  
\usepackage{courier}  
\usepackage[hyphens]{url}  
\usepackage{graphicx} 
\usepackage{natbib}  
\usepackage{caption} 
\usepackage{algorithm}
\usepackage{algorithmic}

\usepackage{newfloat}
\usepackage{listings}
\usepackage{amsthm}
\usepackage{mathtools}
\usepackage{multirow}
\usepackage{amssymb}
\usepackage{pifont}
\usepackage{dsfont}

\usepackage{amsmath}
\usepackage{color}
\usepackage{adjustbox}
\usepackage{colortbl}
\definecolor{Gray}{gray}{0.9}
\usepackage{booktabs}

\DeclareCaptionStyle{ruled}{labelfont=normalfont,labelsep=colon,strut=off} 
\floatstyle{ruled}
\newfloat{listing}{tb}{lst}{}
\floatname{listing}{Listing}
\title{JoVALE: Detecting Human Actions in Video  Using Audiovisual and Language Contexts}
\author{
    Taein Son\textsuperscript{\rm 1}\equalcontrib,
    Soo Won Seo\textsuperscript{\rm 2}\equalcontrib,
    Jisong Kim\textsuperscript{\rm 1},
    Seok Hwan Lee\textsuperscript{\rm 1},
    Jun Won Choi\textsuperscript{\rm 2}\thanks{Corresponding author. Email: junwchoi@snu.ac.kr}
}

\affiliations {
    \textsuperscript{\rm 1}Hanyang University\\
    \textsuperscript{\rm 2}Seoul National University\\
    tison@spa.hanyang.ac.kr, swseo@spa.snu.ac.kr, jskim@spa.hanyang.ac.kr, shlee@spa.hanyang.ac.kr, junwchoi@snu.ac.kr
}

\begin{document}

\maketitle

\begin{abstract}
Video Action Detection (VAD) entails localizing and categorizing action instances within videos, which inherently consist of diverse information sources such as audio, visual cues, and surrounding scene contexts. Leveraging this multi-modal information effectively for VAD poses a significant challenge, as the model must identify action-relevant cues with precision. In this study, we introduce a novel multi-modal VAD architecture, referred to as the Joint Actor-centric Visual, Audio, Language Encoder (JoVALE). JoVALE is the first VAD method to integrate audio and visual features with scene descriptive context sourced from large-capacity image captioning models. At the heart of JoVALE is the actor-centric aggregation of audio, visual, and scene descriptive information, enabling adaptive integration of crucial features for recognizing each actor's actions. We have developed a Transformer-based architecture, the Actor-centric Multi-modal Fusion Network, specifically designed to capture the dynamic interactions among actors and their multi-modal contexts. Our evaluation on three prominent VAD benchmarks—AVA, UCF101-24, and JHMDB51-21—demonstrates that incorporating multi-modal information significantly enhances performance, setting new state-of-the-art performances in the field.
\end{abstract}

\begin{links}
\link{Code}{https://github.com/taeiin/AAAI2025-JoVALE}
\end{links}

\section{Introduction}
Video action detection (VAD) is a challenging task that aims to localize and classify human actions within video sequences. VAD generates bounding boxes with action scores for a keyframe by analyzing the sequential frames around the keyframe. This task differs from {\it Action Recognition} task, which classifies the action for a given video clip, and from {\it Temporal Action Detection} task, which identifies the intervals of particular actions within a video clip.

Humans rely on various sources of information to detect actions, including visual appearance, motion sequences, actor postures, and interactions with their environment.
Numerous studies have demonstrated that leveraging such multi-modal information can significantly enhance action recognition performance \cite{kazakos2019epic, gao2020listen, xiao2020audiovisual, nagrani2021attention}. Audio, in particular, offers valuable information, providing both direct and indirect contextual cues for action recognition. For example, sounds directly linked to actions, like speech, gunshots, or music, can help identify corresponding actions. Additionally, environmental sounds can indirectly suggest relevant actions, such as the sound of waves indicating beach-related activities. Therefore, incorporating audio data alongside visual data can improve the performance and robustness of VAD. Several action recognition methods have successfully utilized both audio and visual data \cite{gao2020listen, xiao2020audiovisual, nagrani2021attention}.

While multi-modal information has shown promise for action recognition tasks, its application in VAD presents significant challenges. Action instances in videos are dispersed across both temporal and spatial dimensions, and the contextual cues necessary for their detection are similarly spread throughout the video. It is crucial to accurately link these actions and contextual features to ensure robust VAD performance. For example, the sound of a piano might help identify a `playing piano' action but would be irrelevant for detecting a `talking with others' action within the same scene. Therefore, the piano sound should be selectively used to detect the `playing piano' action and not the `talking with others' action. Moreover, effectively integrating multi-modal information from various sources is another key to enhancing VAD performance. Despite its potential, the use of audio-visual information for VAD is still relatively unexplored in current research.

\begin{figure*}[!t]
\centering
\includegraphics[width=0.99\textwidth]{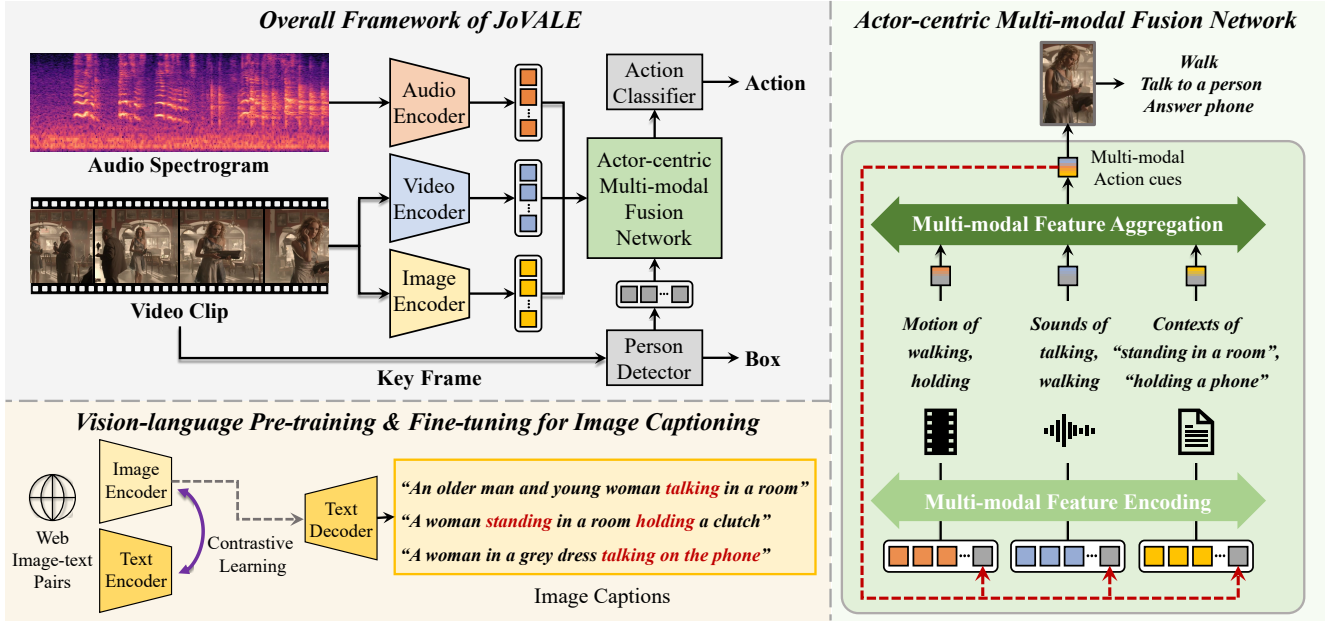}

\caption{Overview of JoVALE: (top-left) The proposed JoVALE integrates audio, visual, and scene-descriptive features using an AMFN.  (bottom-left) JoVALE leverages a VLP model fine-tuned on an image captioning task to generate scene-descriptive features.  (right) AMFN encodes high-level interactions between multi-modal features through MFE and MFA.  }  

\label{fig:fig1}
\end{figure*}

Another valuable resource for VAD identified in this study is the prior general scene-descriptive knowledge gained through vision-language foundation models. Vision Language Pre-training (VLP) models have shown substantial success by leveraging extensive multi-modal data sourced from the web, public databases, and various corpora. These models excel in capturing complex relational structures between text and images, enabling them to adapt to a variety of downstream tasks in either a zero-shot or one-shot manner. With their ability to understand images, the features derived from VLP can significantly enhance VAD performances. In this research, we further investigate a VAD approach that capitalizes on the rich language context provided by VLP models.

This paper introduces a novel multi-modal VAD approach referred to as the Joint Actor-centric Visual, Audio, Language Encoder (JoVALE). JoVALE is the first method to leverage audio and visual modalities alongside language context to localize and classify actions in videos. At the core of JoVALE is the actor-centric modeling of multi-modal contextual information. 

The key concepts of JoVALE are illustrated in Fig. \ref{fig:fig1}. JoVALE begins by generating densely sampled actor proposal features using an off-the-shelf person detector. These actor proposal features are then processed by the Actor-centric Multi-modal Fusion Network (AMFN), which aggregates relevant contextual information from both audio and visual features. Furthermore, AMFN integrates scene-descriptive knowledge acquired from the VLP model, BLIP \cite{li2022blip}, to enrich the action representation.

To fully leverage multi-modal information for VAD, JoVALE effectively models the relationships among actors, temporal dynamics, and various modalities through  AMFN. The AMFN captures their complex interactions through successive updates of Action Embeddings across multiple Transformer layers. It comprises two main components: Multi-modal Feature Encoding (MFE) and Multi-modal Feature Aggregation (MFA).
The MFE module jointly encodes Action Embeddings and Multi-modal Context Embeddings for each modality, achieving computational efficiency through the use of Temporal Bottleneck Features. These Temporal Bottleneck Features provide a compact representation of the temporal changes across all actors. Following this, the MFA module aggregates the Action Embeddings from each modality in a weighted fashion. The resulting features are fed into the subsequent Transformer layer for final action detection.

We evaluated JoVALE on three popular VAD benchmarks: AVA \cite{gu2018ava}, UCF101-24 \cite{soomro2012ucf101}, and JHMDB51-21 \cite{jhuang2013towards}. 
By effectively combining audio, visual, and scene-descriptive context information, JoVALE significantly outperforms the baseline on these benchmarks. On the challenging AVA dataset, JoVALE records a mean Average Precision (mAP) of 40.1\%, achieving a substantial improvement of 2.4\% over the previous best method, EVAD \cite{chen2023efficient}.

Our contributions can be summarized as follows:
\begin{itemize}
\item We present a simple yet effective multi-modal VAD architecture that utilizes the audio-visual information present in videos. Our main approach is Actor-Centric Feature Aggregation, which adaptively attends to the multi-modal context essential for detecting each action instance. There are only a few studies that have explored the use of audio-visual context for VAD. 

\item We are the first to introduce a VAD approach that incorporates general scene-descriptive knowledge inferred from a Vision Language Foundation model. 
\item We propose an efficient architecture that effectively models complex relationships among actors, temporal dynamics, and modalities. Our modeling approach differs from existing VAD methods, which typically combine semantic actor features or predicted scores from each modality in a straightforward manner.  
\end{itemize}

\section{Related Work}

\subsection{Video Action Detection}

Various VAD methods have been proposed, which can be broadly classified into two main approaches: end-to-end and two-stage methods. End-to-end methods predict both the action location and class simultaneously within a single network. These approaches often utilize a Transformer \cite{vaswani2017attention} to predict the set of actions present in a scene. Notable examples of these end-to-end VAD methods include VTr \cite{girdhar2019video}, TubeR \cite{zhao2022tuber}, STMixer \cite{wu2023stmixer}, and EVAD \cite{chen2023efficient}

In contrast, two-stage VAD methods first utilize a pre-trained person detector to localize the actors  before classifying the actions. These two-stage VAD methods include AIA \cite{tang2020asynchronous}, ACAR \cite{pan2021actor}, and JARViS \cite{lee2024jarvis}. Recently, Vision Transformers \cite{tong2022videomae, wang2023videomae, wang2023masked}, pre-trained with Masked Autoencoders (MAE) \cite{he2022masked}, have shown excellent performance in the context of two-stage VAD.

\subsection{Muti-modal Video Action Detection}

Early multi-modal VAD methods \cite{gkioxari2015finding, saha2016deep, zhao2019dance} leveraged both RGB and optical flow to capture appearance and motion information. Another research direction focused on utilizing human skeletal structures through pose estimation models. For example, JMRN \cite{shah2022pose} extracted individual joint features and captured inter-joint correlations. More recently, HIT \cite{faure2023holistic} employed cross-attention mechanisms to capture interactions between key action-related components such as hands, objects, and poses.

Although various video classification methods have utilized both audio and visual information \cite{gao2020listen, xiao2020audiovisual, nagrani2021attention, gong2022contrastive, georgescu2023audiovisual, huang2024mavil}, the application of multi-modal information for VAD has not been thoroughly explored. VAD poses unique challenges, as the relevant audio-visual context needed for accurate detection can vary depending on the specific action instance. This study aims to address this gap.

\section{JoVALE Method}
\label{approach}

\subsection{Overview}

Fig. \ref{fig:fig1} illustrates the overall structure of JoVALE. The model takes audio samples and image frames as input. Audio and visual backbone features  are extracted from these inputs, while scene-descriptive features  are obtained using the BLIP image encoder, pre-trained on an image captioning task. Together, these features form the Multi-modal Embeddings $f_{a}^{(l)}$, $f_{v}^{(l)}$, and $f_{s}^{(l)}$ used for the VAD architecture.

\begin{figure}[t]
\centering
\includegraphics[width=0.42\textwidth]{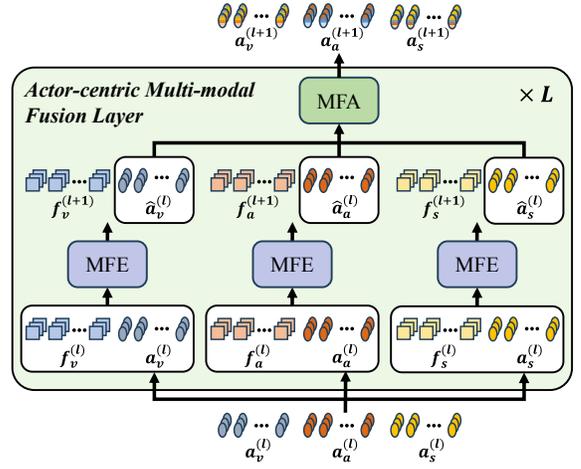}
\caption{Structure of AMFN: Three independent MFEs encode the context features within each modality. Then, MFA combines Action Embeddings derived from each modality.}
\label{fig:fig2}
\end{figure}

JoVALE detects actions through the following steps.
Using the keyframe image, an off-the-shelf person detector generates $K$ actor proposals along with their corresponding Region of Interest (RoI) features, referred to as Actor Proposal Features. The AMFN employs a Transformer to aggregate action-related information from the separately encoded Multi-modal Embeddings $f_{a}^{(l)}$, $f_{v}^{(l)}$, and $f_{s}^{(l)}$. To achieve this, the AMFN jointly encodes the action queries $a_{a}^{(l)}$, $a_{v}^{(l)}$, and $a_{s}^{(l)}$ associated with audio, visual, and scene-descriptive modalities, respectively, where $l$ indicates the layer index. In the first layer, these action queries are all initialized using the Actor Proposal Features.

The structure of AMFN is depicted in Fig.\ref{fig:fig2}. The AMFN comprises two modules: MFE and MFA.  
 For each modality, the MFE module jointly encodes the Action Embeddings, $a_{mod}^{(l)}$ and the Multi-modal Embeddings, $f_{mod}^{(l)}$, producing updated representations $\hat{a}_{mod}^{(l)}$ and $f_{mod}^{(l+1)}$, where $mod \in \mathcal{M} = \{a, v, s\}$.
Subsequently, the MFA module employs an adaptive gated fusion mechanism \cite{kim2018robust} to perform a weighted combination of the three Action Embeddings, $\hat{a}_{a}^{(l)}$, $\hat{a}_{v}^{(l)}$, and $\hat{a}_{s}^{(l)}$, resulting in the combined Action Embeddings $a_{a}^{(l+1)}$, $a_{v}^{(l+1)}$, and $a_{s}^{(l+1)}$. These embeddings are then propagated to the next layer.
This process is repeated over $L$ iterations, progressively refining the Action Embeddings. Finally, the refined Action Embeddings from the $L$-th layer are input into a classifier to predict the action instances.

\subsection{Generation of Multi-modal Features}
\label{MMFE}

\subsubsection{Visual Embeddings.}
We encode an input video clip using a video backbone network such as SlowFast \cite{feichtenhofer2019slowfast} or ViT \cite{dosovitskiy2020image}. This process generates the spatio-temporal visual features $F_{v} \in \mathbb{R}^{T_{v} \times H \times W \times C}$, where $T_v$, $H$, $W$, and $C$ represent temporal, height, width, and channel dimensions, respectively. These features are then reshaped into the visual embeddings $f_{v} \in \mathbb{R}^{T_{v} \times N_{v} \times D}$, where $N_v=HW$ and $D$ denotes the embedding dimension.
\subsubsection{Audio Embeddings.}

Following existing audio preprocessing techniques \cite{gong2021ast}, we transform the audio waveform samples into a log-mel-spectrogram in time and frequency bins.
This spectrogram is fed into a convolution layer with kernel size $P \times P$ and stride $S$, then reshaped into a temporal sequence of $N_a$ feature vectors. This process results in audio embeddings $f_{a} \in \mathbb{R}^{T_{a} \times N_{a} \times D}$.

\begin{figure}[t]
\centering
\includegraphics[width=0.4\textwidth]{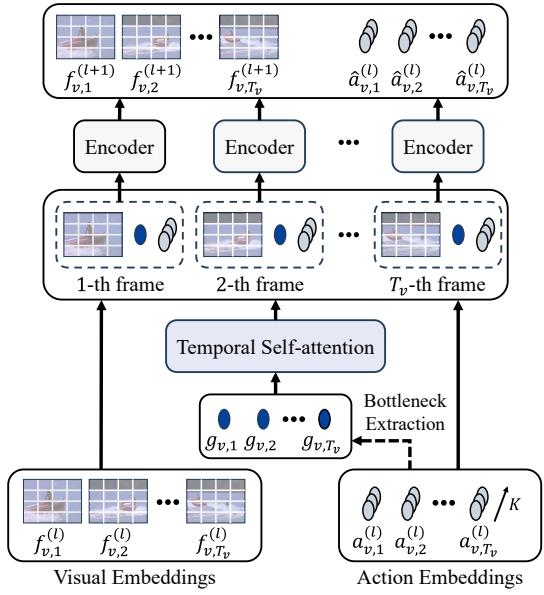}

\caption{Structure of Multi-modal Feature Encoding. This illustration depicts the process for the visual modality. Identical structures are applied individually to other modalities.}  
\label{fig:fig3}
\end{figure}

\subsubsection{Scene-Descriptive Embeddings.}
Scene-descriptive features are generated using the BLIP captioner \cite{li2022blip}, a vision-language foundation model that is fine-tuned on an image captioning task. The BLIP captioner takes each image frame as input, encodes it with an image encoder, and produces a text description of the image through a text decoder. Since the output of the image encoder contains high-level semantic scene information that can be readily translated into the text, we can use it as scene-descriptive features.
We first uniformly sample $T_{s}$ image frames from a video clip. Then, we apply the image encoder of the BLIP captioner to each of $T_{s}$ image frames. The resulting feature maps are then linearly projected into the scene-descriptive embeddings $f_{s} \in \mathbb{R}^{T_{s} \times N_{s} \times D}$.

\subsection{Actor-centric Multi-modal Fusion Network}
\label{ACMM}

AMFN updates Action Embeddings $a_{a}^{(l)}$, $a_{v}^{(l)}$, and $a_{s}^{(l)}$ by applying MFE and MFA in an iterative fashion. 

\begin{table*}[t!]
    \centering
    \fontsize{10}{11}\selectfont
    \setlength{\tabcolsep}{10pt} 
    \begin{tabular}{lcccccc}
        \toprule
        \textbf{Model} & \textbf{Input} & \textbf{Backbone} & \textbf{Pre-train} & \textbf{Val mAP}  \\
         
        \midrule
        \rowcolor{Gray}
        \multicolumn{4}{l}{\textbf{Models with 3D-CNN backbones}} & \quad & \quad \\
        WOO \cite{chen2021watch} & 32 $\times$ 2 & SF-R101 & K600 & 28.3 \\
        SlowFast \cite{feichtenhofer2019slowfast} & 32 $\times$ 2 & SF-R101 & K600 & 29.0 \\
        AIA \cite{tang2020asynchronous} & 32 $\times$ 2 & SF-R101 & K700 & 32.3 \\
        ACAR \cite{pan2021actor} & 32 $\times$ 2 & SF-R101 & K700 & 33.3 \\
        TubeR \cite{zhao2022tuber} & 32 $\times$ 2 & CSN-152 & K400 & 33.6 \\
        HIT \cite{faure2023holistic} & 32 $\times$ 2 & SF-R101 & K700 & 32.6 \\
        STMixer \cite{wu2023stmixer} & 32 $\times$ 2 & SF-R101 & K700 & 30.9 \\
        
        \midrule
        
        \textbf{JoVALE} & 32 $\times$ 2 & SF-R101 & K700 & \textbf{35.5} \\ 
        
        \midrule
        \rowcolor{Gray}
        \multicolumn{4}{l}{\textbf{Models with ViT backbones}} & \quad & \quad \\
        VideoMAE \cite{tong2022videomae} & 16 $\times$ 4 & ViT-B & K400 & 31.8 \\
        MViTv2 \cite{li2022mvitv2} & 32 $\times$ 3 & MViTv2-B & K700 & 32.3 \\
        MeMViT \cite{wu2022memvit} & 32 $\times$ 3 & MViTv2-B & K700 & 34.4 \\
        MVD \cite{wang2023masked} & 16 $\times$ 4 & ViT-B & K400 & 34.2 \\
        STMixer \cite{wu2023stmixer} & 16 $\times$ 4 & ViT-B{\scriptsize $^\dagger$} & K710 & 36.1 \\
        EVAD \cite{chen2023efficient} & 16 $\times$ 4 & ViT-B{\scriptsize $^\dagger$} & K710 & 37.7 \\
        
        \midrule
        
        \textbf{JoVALE} & 16 $\times$ 4 & ViT-B{\scriptsize $^\dagger$} & K710 & \textbf{40.1} \\
        
        \bottomrule
    
    \end{tabular}
    \caption{Performance comparison evaluated on the AVA 2.2 dataset. ViT-B marked with {\scriptsize $^\dagger$} is initialized with pre-trained weights from VideoMAE v2 \cite{wang2023videomae}.}
    \label{table_ava}
\end{table*}

\subsubsection{Multi-modal Feature Encoding.}
\label{MFE}
The structure of MFE is shown in Fig.~\ref{fig:fig3}.
The  MFE performs the following operation 
\begin{align}
    {\hat{a}}_{mod}^{(l)}, f_{mod}^{(l+1)}  &= \text{MFE}\left(  a_{mod}^{(l)} , f_{mod}^{(l)}\right).
   \end{align}
Applying self-attention to the combination of $a_{mod}^{(l)}$ and $f_{mod}^{(l)}$ can result in high computational complexity, especially when the number of  embeddings is large. To address this, we generate Temporal Bottleneck Features, which compress the input embeddings across actors at each time step, effectively reducing the computational overhead.
By taking $a_{mod}^{(l)} \in \mathbb{R}^{K \times T_{mod} \times D}$ as an input, MFE computes the Temporal Bottleneck Features, $b_{mod}^{(l)} \in \mathbb{R}^{T _{mod} \times D}$ from
\begin{align}
b_{mod}^{(l)} = \text{SA}(\text{Pool}(a_{mod}^{(l)})),
\end{align}
where Pool refers to the average pooling over the actor dimension, and SA denotes the multi-head self-attention. Note that this SA operation encodes the Action Embeddings in the time domain. 
Finally, the Temporal Bottleneck Features are merged into the Multi-modal Embeddings $f_{{mod},t}^{(l)}$ and the Action Embeddings $a_{{mod},t}^{(l)}$. Then, MFE jointly encodes the merged embeddings for each time step
\begin{align}
    \hat{a}_{{mod},t}^{(l)}, f_{{mod},t}^{(l+1)}  = \text{Encoder}([a_{{mod},t}^{(l)},f_{{mod},t}^{(l)},b_{{mod},t}^{(l)}]),
\end{align}
where $t\in[1,T_{mod}]$, Encoder consists of a SA, two normalization layers, and an FFN. Finally, the updated Action Embeddings  $\hat{a}_{{a}}^{(l)}$, $\hat{a}_{{v}}^{(l)}$, and $\hat{a}_{{s}}^{(l)}$ are delivered to MFA module. 

\begin{figure}[!t]
\centering
\includegraphics[width=0.42\textwidth]{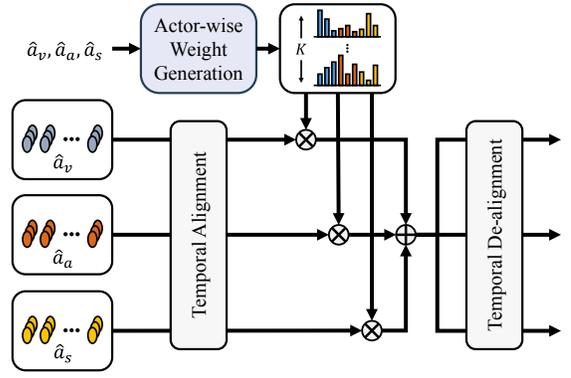}
\caption{Structure of Multi-modal Feature Aggregation.}  
\label{fig:fig4}
\end{figure}
\subsubsection{Multi-modal Feature Aggregation.}
\label{MFA}
The structure of MFA is depicted in Fig.~\ref{fig:fig4}.
The MFA operates as follows
\begin{align}
      {a}_v^{(l+1)}, {a}_a^{(l+1)}, {a}_s^{(l+1)}=\text{MFA} ({{\hat{a}}_v}^{(l)},{{\hat{a}}_a}^{(l)},{{\hat{a}}_s}^{(l)}).
\end{align}
MFA starts with Temporal Alignment, which aligns time sampling between the Action Embeddings  ${{\hat{a}}_v}^{(l)}$, ${{\hat{a}}_a}^{(l)}$, and ${{\hat{a}}_s}^{(l)}$.
 Following \cite{cooper2019hear},  the Action Embeddings of size $T_{mod}$ in time dimension are resized to those of the fixed size $T_c$.
Then, MFA adaptively integrates query features ${{\hat{a}}_v}^{(l)},{{\hat{a}}_a}^{(l)},{{\hat{a}}_s}^{(l)}$ using an adaptive gated fusion mechanism \cite{kim2018robust}.
We compute the combining weights $w_v^{(l)}$,  $w_a^{(l)}$, and  $w_s^{(l)}$ for each actor.  We first concatenate Action Embeddings ${{\hat{a}}_v}^{(l)},{{\hat{a}}_a}^{(l)},{{\hat{a}}_s}^{(l)}$ and applies average pooling over the time dimension
\begin{align}
a_p^{(l)}=\text{Pool}([\hat{a}_{v}^{(l)}\parallel\hat{a}_{a}^{(l)}\parallel\hat{a}_{s}^{(l)}]),
\end{align}
where $\parallel$ denotes channel-wise concatenation. 
Then, we obtain the combining weights by passing the pooled features through a bottleneck MLP followed by a sigmoid function 
\begin{align}
[w_v^{(l)}\parallel w_a^{(l)}\parallel w_s^{(l)}]=\sigma(\text{MLP}(a_p^{(l)})),
\end{align}
where MLP consists of two fully connected layers with an activation function and $\sigma(\cdot)$ denotes sigmoid function. Then, the updated Action Embeddings $a_{a}^{(l+1)},a_{v}^{(l+1)}$, and $a_{s}^{(l+1)}$ are obtained by a weighted summation
\begin{align}
a_{fuse}^{(l)}&=\frac{1}{|\mathcal{M}|} \sum_{m \in \mathcal{M}} w_{m}^{(l)} \otimes \hat{a}_{m}^{(l)}\\
a_{mod}^{(l+1)}&=a_{fuse}^{(l)}+ w_{mod}^{(l)} \otimes \hat{a}_{mod}^{(l)},
\end{align}
where $\otimes$ denotes the element-wise multiplication. 

The updated Action Embeddings $a_{a}^{(l+1)},a_{v}^{(l+1)},a_{s}^{(l+1)}$ are converted back to those of their own temporal sampling. After $L$ layers, the classification head is applied to ${a_{fuse}^{(L)}}$ to predict action scores $c \in \mathbb{R}^{K \times N_{\text{cls}}}$, where $N_{\text{cls}}$ denotes the number of target classes and $K$ is the number of actor proposals. These scores, along with the corresponding actor bounding boxes $b \in \mathbb{R}^{K \times 4}$,  form a set of action instances $(b, c)$.

\section{Experiments}
\subsection{Datasets and Metrics}

We evaluate JoVALE on three standard VAD datasets: AVA \cite{gu2018ava}, UCF101-24 \cite{soomro2012ucf101}, and JHMDB51-21 \cite{jhuang2013towards}. AVA consists of 299 15-minute movie clips, with 235 for training and 64 for validation. We evaluate our approach on 60 action classes in AVA v2.2. 

UCF101-24, a subset of UCF101, contains 24 sport action classes with 3,207 instances, and our method is evaluated on the first split. 

JHMDB51-21, a subset of JHMDB51, includes 928 trimmed video clips spanning 21 action classes. 

We report the average performance across the three standard splits of the dataset. The evaluation metric is frame-level mAP at an Intersection over Union (IoU) threshold of 0.5 for all datasets.

\subsection{Implementation Details}
In this section, we describe the implementation details of our proposed JoVALE. 

\subsubsection{Hyperparameters.}
The AMFN consists of $L=6$ Transformer layers. When conducting temporal alignment and de-alignment within the MFA module, the temporal dimension $T_c$
  is aligned to match that of the visual images, $T_v$. Following the approach in \cite{cooper2019hear}, we apply temporal average pooling when $T_{mod} \geq T_c$ and temporal repetition when $T_{mod} < T_c$.
The temporal length of the visual features $T_v$ varies based on the chosen backbone architecture. Specifically, $T_v$ is set to 4 when utilizing the SlowFast \cite{feichtenhofer2019slowfast} architecture and 8 with the ViT \cite{dosovitskiy2020image}.
For audio data, the spectrograms are processed through a convolutional layer with a kernel size of $P = 16$ and a stride of $S = 10$, yielding audio embeddings with a temporal length of $T_a = 20$.
Regarding the scene-descriptive features, the number of input frames inputted into BLIP is set at $T_s=4$. 
The hyperparameter $D$, representing the Transformer embedding size, is set to 256.

\subsubsection{Generation of Multi-modal Features.}
\begin{table}[t!]
\centering
\setlength{\tabcolsep}{1mm}
\fontsize{9}{10}\selectfont
\begin{tabular}{lcccccc}
    \toprule
    Model                           &Input               & Backbone  & Pre-train  & UCF &JHMDB \\
    \midrule
    AVA             &20 $\times$ 1     & I3D  & K400        & 76.3 & 73.3               \\
    AIA  &32 $\times$ 1       & C2D  & K400                 & 78.8 & -               \\
    ACRN         &20 $\times$ 1     & S3D-G  & K400         & -    & 77.9 \\
    CARN        &32 $\times$ 2     & I3D-R50  & K400        & -    & 79.2 \\
    YOWO       &16 $\times$ 1       & 3D-RX-101 &K400       & 75.7 & 80.4           \\
    WOO         &32 $\times$ 2     & SF-R101 & K600         &  -   & 80.5 \\
    TubeR       &32 $\times$ 2       & CSN-152 & K400       & 83.2 &  -           \\
    $\text{ACAR}^{*}$  &32 $\times$ 1   & SF-R50  & K400                 & 84.3 &  -                     \\
    $\text{HIT}^{*}$         &32 $\times$ 2       & SF-R50      & K700   & 84.8 & 83.8           \\                
    STMixer     &32 $\times$ 2       & SF-R101  & K700      & 83.7 & 86.7           \\   
    \midrule
    \textbf{JoVALE}                 &32 $\times$ 2   & SF-R101  &K700 & \textbf{84.9} &\textbf{91.0}       \\
    \bottomrule
    \end{tabular}
\caption{Performance comparison on UCF101-24 and JHMDB51-21. The models marked with $*$ employ YOWO \cite{kopuklu2019you} as a person detector.}
\label{table_ucf_and_jhmdb}
\end{table}
Visual features were extracted using one of the following backbones: 1) SlowFast-R50 pre-trained on Kinetics-400 \cite{kay2017kinetics}, 2) SlowFast-R101 pre-trained on Kinetics-700 \cite{carreira2019short}, or 3) ViT-B with pre-trained weights from VideoMAE v2.

Audio preprocessing followed the approach in \cite{gong2021ast}, where log-mel-spectrograms were extracted from raw audio waveforms. The waveforms, sampled at 16kHz, were converted into 128 Mel-frequency bands using a 25ms Hamming window with a 10ms stride. For an input audio clip of $t$ seconds, this process produced spectrograms of dimension $100t \times 128$.

Scene-descriptive features were extracted using the ViT-B BLIP captioner, which was pre-trained on images from COCO, Visual Genome \cite{krishna2017visual}, and web datasets \cite{changpinyo2021conceptual, ordonez2011im2text, schuhmann2021laion}, and subsequently fine-tuned on the COCO Caption dataset.

\subsubsection{Initializing Action Embeddings.}
We employed a Faster-RCNN \cite{ren2015faster} with a ResNeXt-101-FPN \cite{lin2017feature, xie2017aggregated} as a person detector.
The detector was pre-trained on ImageNet \cite{russakovsky2015imagenet} and COCO human keypoint images \cite{lin2014microsoft} and fine-tuned on each target VAD dataset. 
The top $K=15$ actor features were extracted from the penultimate layer based on human confidence scores.

\subsubsection{Training.}
The pre-trained person detector and image captioner were kept frozen during both training and inference. The entire model was trained using sigmoid focal loss for action classification. The AdamW optimizer was employed with a weight decay of 1e-4. Initial learning rates were set to 1e-5 for the video backbone and 1e-4 for the other networks, with a tenfold reduction applied at the 7th epoch. Training was conducted for 8 epochs with a batch size of 16, utilizing four NVIDIA GeForce RTX 3090 GPUs.

For data augmentation, we applied random horizontal flipping to RGB frames. For audio, we utilized SpecAugment \cite{park2019specaugment} with time and frequency masking, following the approach in AST \cite{gong2021ast}.

\subsection{Main Results}
\subsubsection{Performance Comparison.}

We compare JoVALE against existing VAD methods across three widely used datasets. As shown in Table \ref{table_ava}, on the AVA dataset, JoVALE outperforms all other methods on the AVA dataset when employing both 3D-CNN and ViT backbones. With a 3D-CNN backbone, JoVALE surpasses the previously leading method, TubeR \cite{zhao2022tuber}, by 1.9\% mAP. When utilizing a ViT backbone, JoVALE establishes a new state-of-the-art on AVA, outperforming EVAD \cite{chen2023efficient} by 2.4\% mAP.

The results evaluated on the UCF101-24 and JHMDB51-21 datasets are shown in Table \ref{table_ucf_and_jhmdb}. Here, we used SF-101 as the visual backbone. Given that over 80\% of video clips in these datasets lack audio, JoVALE relies solely on visual and scene-descriptive features for input. Even without audio, JoVALE achieves state-of-the-art performance, with mAP scores of 84.9\% on UCF101-24 and 91.0\% on JHMDB51-21.

\subsubsection{Computational Analysis.}

Table \ref{table_computation} compares the computational costs of JoVALE with those of other methods. 
JoVALE exhibits higher computational complexity compared to other methods, largely due to its incorporation of audio and video data, along with the use of the BLIP model. However, the increased complexity is justified by JoVALE's superior performance and remains within a reasonable range. It has been observed that reducing the input resolution of BLIP can lower computational costs, albeit at the expense of a slight decrease in performance.

\begin{table}[t!]
\centering
\setlength{\tabcolsep}{1mm}
\fontsize{9}{10}\selectfont
\begin{tabular}{lccccc}
\toprule
Model  &  Input   & Backbone   & GFLOPs   & mAP        \\
\midrule
SlowFast               &  $32 \times 2$   & SF-R101-NL       & 199 + NA & 29.0    \\
ACAR                      & $32 \times 2$   & SF-R101          & 160 + NA & 31.7    \\
VideoMAE                     & $16 \times 4$   & ViT-B            & 180 + NA & 31.8    \\
MeMViT                      & $32 \times 3$   & MViTv2-B          & 212 + NA      & 33.5    \\
WOO                          & $32 \times 2$   & SF-R101-NL       & 252      & 28.3    \\
TubeR                      & $32 \times 2$   & CSN-152          & 240      & 32.0    \\
STMixer                    & $16 \times 4$   & ViT-B{\scriptsize $^\dagger$}  & 355      & 36.1    \\
EVAD                         & $16 \times 4$   &ViT-B{\scriptsize $^\dagger$}  & 243      & 37.7    \\
\midrule
\textbf{JoVALE}    & $16 \times 4$   & ViT-B{\scriptsize $^\dagger$}  &  495        &     \textbf{40.1}\\
\textbf{JoVALE}@$288$         & $16 \times 4$   & ViT-B{\scriptsize $^\dagger$}  &  387       &   39.8  \\
\textbf{JoVALE}@$192$         & $16 \times 4$   & ViT-B{\scriptsize $^\dagger$}  &     314    &   39.3  \\
\bottomrule
\end{tabular}
\caption{Computational costs comparison evaluated on AVA 2.2 dataset. $\textbf{JoVALE@}N$ indicates $N\times N$ input size for BLIP. `NA' is the person detector's cost, which was not reported in the corresponding paper. }
\label{table_computation}

\end{table}

\begin{figure*}[!hbt]
\centering
\includegraphics[width=0.99\textwidth]{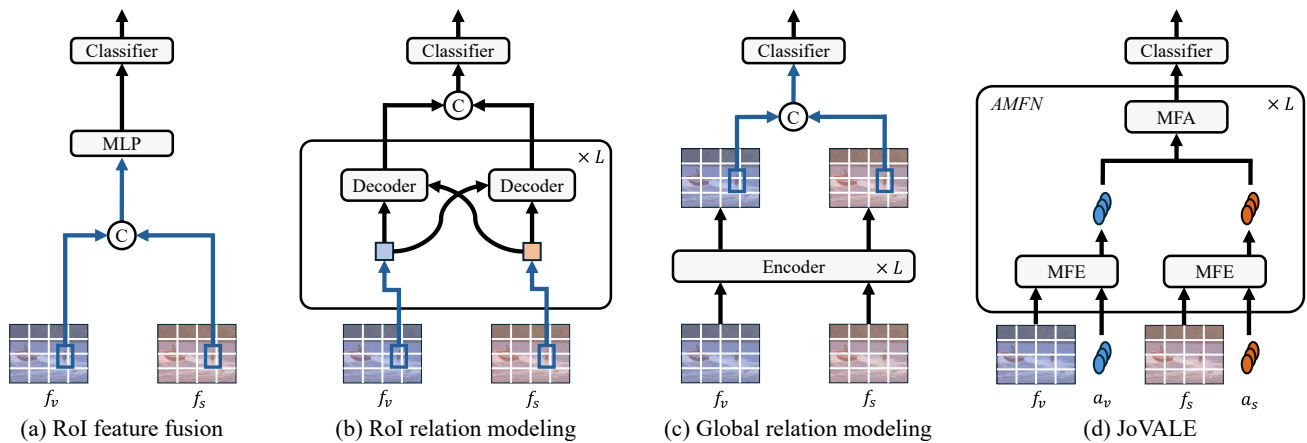}
\caption{Different multi-modal fusion strategies: The symbol \textcircled{c} denotes the channel-wise concatenation.}
\label{fig:fig4_for_tab4}
\end{figure*}

\subsection{Ablation Study}
Our ablation studies were conducted on AVA v2.2 using the SlowFast-R50 configuration. Unless stated otherwise, all other settings were consistent with the main experiments. Detailed model configurations used in these ablations are provided in the {\it Supplementary Material}.

\subsubsection{Multi-modalities.}

In Table \ref{table_abl_0}, we assess the performance of JoVALE using various modality combinations. Using only the video modality, we achieve the highest mAP of 28.0\%, highlighting its essential role in VAD. In contrast, relying solely on audio results in significantly lower performance, illustrating its limitations as an independent modality.

By integrating video with scene descriptive embeddings, a notable improvement to an mAP of 32.7\% is observed. This enhancement underscores the effectiveness of combining visual and scene-descriptive contexts to boost VAD performance. Notably, when all three modalities—audio, video, and scene descriptive—are utilized together, JoVALE achieves the highest mAP of 34.0\%. This underscores the benefits of leveraging the complementary nature of diverse modalities in action detection.





\begin{table}[t]
\centering
\fontsize{9}{10}\selectfont
    \begin{tabular}{l>{\centering\arraybackslash}p{1cm} >{\centering\arraybackslash}p{1cm} >{\centering\arraybackslash}p{1.6
    cm}c}
    \toprule
    \multirow{2}{*}{Method} &\multicolumn{3}{c}{Modality}       & \multirow{2}{*}{mAP}  \\
    \cmidrule(r){2-4}
                    & Video     &  Audio   & Scene-desc.    &        \\
    \midrule
    \multirow{3}{*}{Uni-modal}& \checkmark  &           &              & 28.0   \\
    &           & \checkmark &              & 11.5          \\
    &           &           & \checkmark    & 26.9           \\
    \midrule
    \multirow{4}{*}{Multi-modal} & \checkmark  & \checkmark &              & 28.6           \\
    &\checkmark  &           & \checkmark    & 32.7    \\
    &           & \checkmark & \checkmark    & 27.8    \\    
    &\checkmark  & \checkmark & \checkmark    & \textbf{34.0}    \\
    \bottomrule
\end{tabular}        

\caption{Performance evaluated when various combinations of modalities are used.} 
\label{table_abl_0}

\end{table}

\begin{figure*}[!t]
\centering
\includegraphics[width=0.95\textwidth]{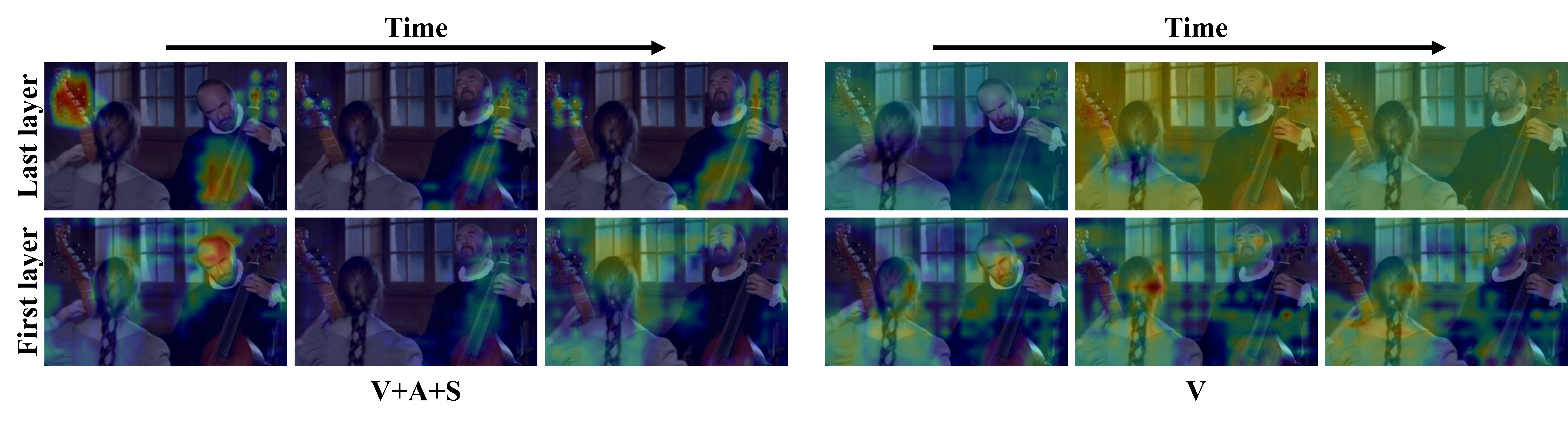}
\caption{Visualization of activation maps: The left side displays heatmaps when using audio, visual, and scene-descriptive features for JoVALE, while the right side shows heatmaps based solely on visual features.}
\label{fig:fig5}
\end{figure*}

\subsubsection{Multi-modal Fusion Strategies.}

In Table \ref{table_abl_1}, we compare the performance of various multi-modal fusion strategies commonly used in VAD, with Fig.~\ref{fig:fig4_for_tab4} illustrating the different strategies considered. Specifically, we evaluate (a) RoI feature fusion \cite{gkioxari2015finding}, (b) RoI relation modeling \cite{faure2023holistic}, and (c) Global relation modeling. For a fair comparison, audio features were excluded from the fusion process.

RoI feature fusion \cite{gkioxari2015finding} directly combines RoI actor features from each modality for action classification. RoI relation modeling \cite{faure2023holistic} uses cross-attention to capture relationships among RoI features from different modalities. In global relation modeling, a transformer encoder is utilized to capture holistic dependencies across multi-modal embeddings.
Notably, the feature fusion strategy employed in JoVALE achieves the best performance among all evaluated strategies.
\begin{table}[t!]
\centering
\fontsize{9}{10}\selectfont
    \begin{tabular}{lcc}
    \toprule
   Methods         & mAP \\
    \midrule
    RoI feature fusion           & 29.1   \\
    RoI relation modeling              & 30.4   \\
    Global relation modeling              &  32.1  \\
    \midrule
    
    \textbf{JoVALE}             & \textbf{32.7}   \\
    \bottomrule
\end{tabular}        

\caption{Performance of different multi-modal fusion strategies.}
\label{table_abl_1}
\end{table}

\subsubsection{MFE Structure.}

We explored various MFE structures for spatio-temporal feature extraction, with results presented in Table \ref{table_abl_2}.
We first established a baseline using joint space-time attention that encodes spatio-temporal features within a single encoder. While achieving 33.8 mAP, this approach incurs substantial computational overhead.
Factorized encoder \cite{arnab2021vivit} first extracts features from individual frames, then captures temporal relationships between them. While this sequential encoding reduces complexity, it yields lower performance at 30.6 mAP. 
Divided space-time attention \cite{bertasius2021space}, which applies temporal and spatial attention separately within each Transformer layer, achieves 33.1 mAP with 31.2 GFLOPs.
Cross-frame attention \cite{ni2022expanding}, which uses randomly initialized tokens for inter-frame information exchange, resulting in a 1.4 mAP decrease compared to the baseline.
In contrast, our MFE leverages bottleneck features derived from actor features, enabling the exchange of crucial actor-centric information. By focusing on actor-relevant information, our method effectively balances the trade-off between model complexity and performance, achieving superior performance with 34.0 mAP and maintaining computational efficiency at 25.4 GFLOPs.

\begin{table}[t!]
\centering
\fontsize{9}{10}\selectfont
    \begin{tabular}{lcc}
    \toprule
    Multi-modal Feature Encoding       & GFLOPs & mAP \\
    \midrule
    Joint space-time attention (Baseline)     &       41.7      & 33.8   \\
    Factorized encoder    &         23.0     & 30.6   \\
    Divided space-time attention    &        31.2     & 33.1   \\
    Cross-frame attention   &        24.4    & 32.4   \\
    \textbf{MFE}         &       25.4     & \textbf{34.0}   \\
    \bottomrule
\end{tabular}        

\caption{Performance of different spatio-temporal feature encoding approaches.}
\label{table_abl_2}
\end{table}

\subsubsection{Effect of Adaptive Gated Fusion in MFA.}
 
Table \ref{table_abl_gate} compares the performance of the model with Adaptive Gated Fusion enabled against other prevalent multi-modal fusion techniques. Initially, we established a baseline using Late score fusion, which achieves an mAP of 29.4. We then experimented with the case where action embeddings are combined using equal weights, yielding an mAP of 31.9. We confirm that Adaptive Gated Fusion provides a 2.1\% mAP improvement over the baseline.

\subsection{Qualitative Results}

Fig.~\ref{fig:fig5} compares the activation maps when using visual, audio, and scene-descriptive features together versus using only visual features. The top row shows activation maps from the first layer, while the bottom row presents maps from the final layer. With multi-modal input, JoVALE demonstrates improved localization of regions of interest compared to using visual input alone. Notably, JoVALE successfully focuses on a cello when both visual and audio data are provided but fails to do so with visual input only. These results highlight the interactions between visual and audio features, which leads to better extraction of visual cues.

\section{Conclusions}
\begin{table}[t!]
\centering
\fontsize{9}{10}\selectfont

    \begin{tabular}{lcc}
    \toprule
   Methods         & mAP \\
    \midrule
    Late score fusion              & 29.4   \\
    Uniform weighted fusion              & 31.9   \\
    \midrule
    \textbf{Adaptive Gated Fusion}             & \textbf{34.0}   \\
    \bottomrule
\end{tabular}        

\caption{Effect of adaptive gated fusion in MFA.}
\label{table_abl_gate}
\end{table}
In this paper, we introduced JoVALE, a multi-modal VAD network that effectively extracts audio, visual, and scene-descriptive contexts from the input. JoVALE selectively integrates critical information from each modality to detect various actions within a scene. Built on a Transformer architecture, JoVALE attends to features from each modality using actor features, identified by a person detector, as queries. The AMFN module facilitates computationally efficient modeling of high-level relationships among actors and the temporal dynamics across different modalities. It jointly encodes visual, audio, and scene-descriptive embeddings through the MFE and aggregates them with adaptive weights for each actor through the MFA. Evaluations on challenging VAD benchmarks demonstrate that JoVALE achieves state-of-the-art performance, significantly outperforming existing VAD methods by notable margins.

While we have utilized fine-tuned image captioning models to extract scene-context information, exploring the potential of generic pre-trained VLMs to further enhance VAD is an exciting direction. These models could offer a high-level understanding of a scene, which may significantly boost VAD performance. This enhancement could be achieved by designing effective prompting strategies and integrating tokens generated by foundation models into VAD architectures. We plan to pursue this line of research in future work.

\section{Acknowledgements}
This work was supported by Institute of Information \& communications Technology Planning \& Evaluation (IITP) grant funded by the Korea government(MSIT) [NO.RS-2021-II211343, Artificial Intelligence Graduate School Program (Seoul National University)], the National Research Foundation of Korea(NRF) grant funded by the Korea government(MSIT) (No.2020R1A2C2012146), and the Technology Innovation Program (RS-2024-00468747, Development of AI and Lightweight Technology for Embedding Multisensory Intelligence Modules) funded By the Ministry of Trade Industry \& Energy (MOTIE, Korea).

\bibliography{jovale}

\end{document}